\pdfoutput=1

\documentclass[11pt]{article}

\usepackage{acl}

\usepackage{times}
\usepackage{latexsym}

\usepackage[T1]{fontenc}
\usepackage[utf8]{inputenc}

\usepackage{microtype}

\usepackage{algorithm}
\usepackage{algorithmic}
\usepackage{mathrsfs}
\usepackage{amsmath}
\usepackage{graphicx}
\usepackage{booktabs}
\usepackage{array}
\usepackage{boldline}
\usepackage{multirow,tabularx}
\usepackage{diagbox}
\usepackage{enumitem}
\setitemize{noitemsep,topsep=0pt,parsep=0pt,partopsep=0pt}
\setenumerate{noitemsep,topsep=0pt,parsep=0pt,partopsep=0pt}

\title{Database Search Results Disambiguation for Task-Oriented Dialog Systems}

\author{Kun Qian$^{\dagger}$\thanks{\hspace{0.15cm}Work done during KQ`s internship at Meta AI}, Satwik Kottur$^{\ddagger}$, Ahmad Beirami$^{\ddagger}$,  Shahin Shayandeh$^{\ddagger}$, Paul Crook$^{\ddagger}$, \\
\textbf{Alborz Geramifard$^{\ddagger}$, }\textbf{Zhou Yu}$^{\dagger}$\textbf{, Chinnadhurai Sankar}$^{\ddagger}$ \\
  $^{\dagger}$Columbia University \\
  \texttt{\{kq2157, zy2461\}@columbia.edu} \\
  $^{\ddagger}$Meta AI\\
  \texttt{\{beirami, skottur, shn, pacrook, alborzg, chinnadhurai\}@fb.com}}
  
\begin{document}
\maketitle
\begin{abstract}
As task-oriented dialog systems are becoming increasingly popular in our lives, more realistic tasks have been proposed and explored.
However, new practical challenges arise. 
For instance, current dialog systems cannot effectively handle multiple
search results when querying a database, due to lack of such scenarios in 
existing public datasets.
In this paper, we propose \textit{Database Search Result (DSR) Disambiguation}, a novel task that focuses on disambiguating database search results, which enhances user experience by allowing them to choose from multiple options instead of just one.
To study this task, we augment the popular task-oriented dialog datasets (MultiWOZ and SGD) with turns that resolve ambiguities by
(a) synthetically generating turns through a pre-defined grammar, and
(b) collecting human paraphrases for a subset.
We find that training on our augmented dialog data improves the model's ability to deal with ambiguous scenarios, without sacrificing performance on unmodified turns. 
Furthermore, pre-fine tuning and multi-task learning helps our model to improve performance on DSR-disambiguation even in the absence of in-domain data, suggesting that it can be learned as a universal dialog skill. 
Our data and code will be made publicly available.
\end{abstract}

\section{Introduction}
\begin{figure}[t]
\centering
\includegraphics[width=0.97\columnwidth]{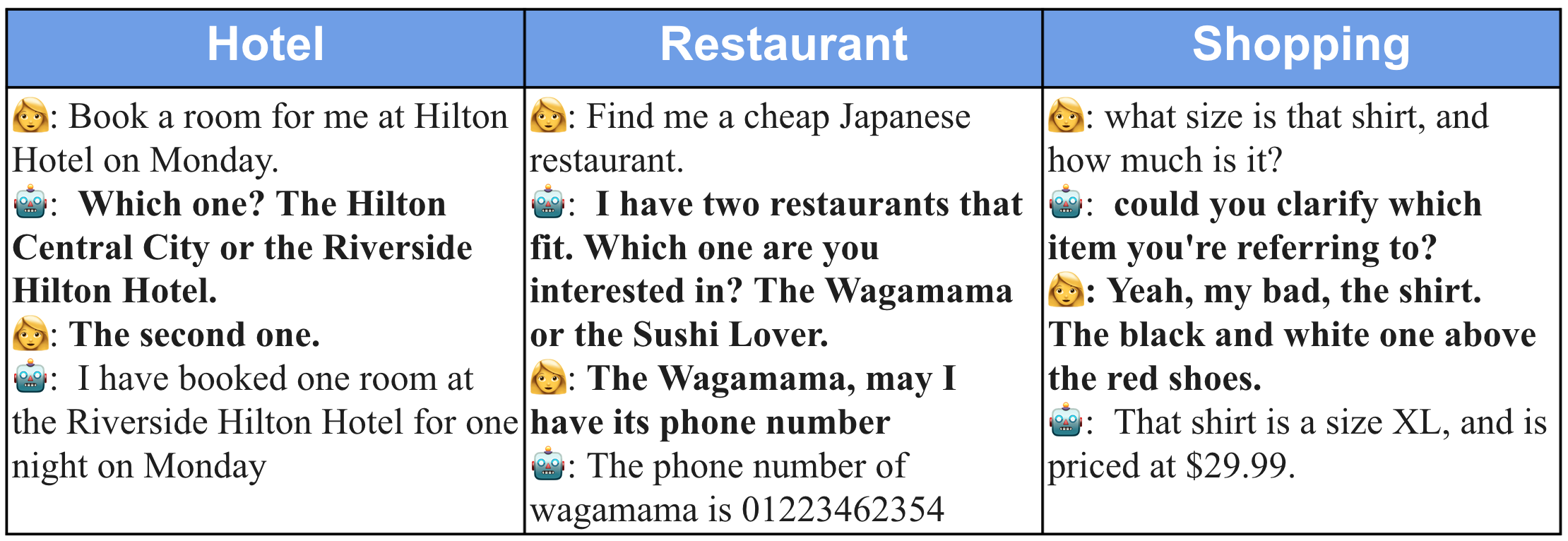}
\caption{Examples of disambiguation turns over three different domains. 
}
\label{fig_case}
\end{figure}

Task-oriented dialog (TOD) systems have been widely deployed for popular virtual assistants, like Siri and Google Assistant.
They help people with tasks such as booking restaurants and looking for a hotel by searching databases with constraints provided by users. 
After retrieving a result from the database, a system may continue by conducting actions like making a reservation or providing more information about receiving the result.
However, there can be multiple results from the database that match the same constraints. 
For example, as shown in Fig.~\ref{fig_case}, the system finds two available hotels at different locations when the user is asking the system to help book a hotel. 
This kind of ambiguity stops system from proceeding until the system finds out which result the user looks for.
Therefore, we need to enhance the system with the ability to resolve such ambiguity brought out by multiple items returned from database search.
We call this type of ambiguity as database search result ambiguity (DSR-ambiguity).

Different from semantic ambiguous words (e.g. ``orange'' can be referred as either color or fruit), the DSR-ambiguity focuses on results from multiple database search results. Solving such disambiguation tasks  consists of two steps: asking clarification questions and understanding user's corresponding answers. 
While there is a relatively larger body of literature focusing on when and how to give out the clarification question~\cite{rao-daume-iii-2018-learning, Rao2019AnswerbasedAT, kumar-black-2020-clarq}, the focus on understanding user's answers/intents has been relatively sparse.
Our work mainly focuses on improving model's ability of understanding the answers by augmenting two existing task-oriented dialog datasets: MultiWOZ~\cite{budzianowski-etal-2018-multiwoz} and Schema-Guided Dataset (SGD)~\cite{rastogi2019towards}.

MultiWOZ and SGD are the most popular large-scale task-oriented dialog datasets, based on which most of the state-of-the-art dialog system models are commonly trained and evaluated.
According to our analysis, there are around 66\% dialogs of the dataset contains multiple dataset-searching results, which means the DSR-ambiguity exists.
In this setting, ambiguities are skipped and the model trained based on these datasets can hardly handle the cases where users prefer to make their own choices among all the results satisfies the constraints.
Furthermore, users should be given more detailed information about search results.
Ideally, dialog models should provide the information and assist users to make choices, rather than picking one from the result list and recommending it to users. 
It is not necessary to list all the results, but enumerating 2 or 3 options would help increase user's engagement.
To strengthen the model with the ability to handle the ambiguity, we propose to augment these two datasets with disambiguation turns, where the system provides all possible matched results and lets the user make their own decision based on the complete information.

Specifically, we first extract templates from the SIMMC 2.0 dataset~\cite{Kottur2021SIMMC2A}, which is a multi-modal task-oriented dialog dataset containing disambiguation turns but only covering two domains. 
Based on the extracted templates and database from MultiWOZ and SGD, we synthesize a one-turn dialog dataset, containing only the disambiguation turn, to check whether the model can learn the disambiguation from the data.
To be applicable in reality, we expect the model to learn the skill of disambiguation without compromising the performance on other dialog skills. 
So, we propose to augment the MultiWOZ and SGD with disambiguation turns and train dialog models with the augmented dataset.
To ensure naturalness and diversity of the automatically augmented dataset, we additionally recruit crowd-workers to paraphrase the modified turns.

In conclusion, our contribution includes:
\begin{enumerate}
    \item We propose \textit{Database Search Result Disambiguation}, a new dialog task
    focused on understanding the user's needs through clarification questions.
    \item We provide a generic framework for augmenting disambiguation turns, and apply this framework to augment the two most popular task-oriented dialog datasets with disambiguation cases. We also conduct human paraphrasing for the augmented utterances in test sets.
    \item We create a benchmark for the new task with pre-trained GPT2 model. The results show that our augmented dataset enhances the model's disambiguation ability, while maintaining the performance on the original tasks.
\end{enumerate}

\section{Task Formulation}
\label{sec:task_form}
\begin{figure}[t]
\centering
\includegraphics[width=0.97\columnwidth]{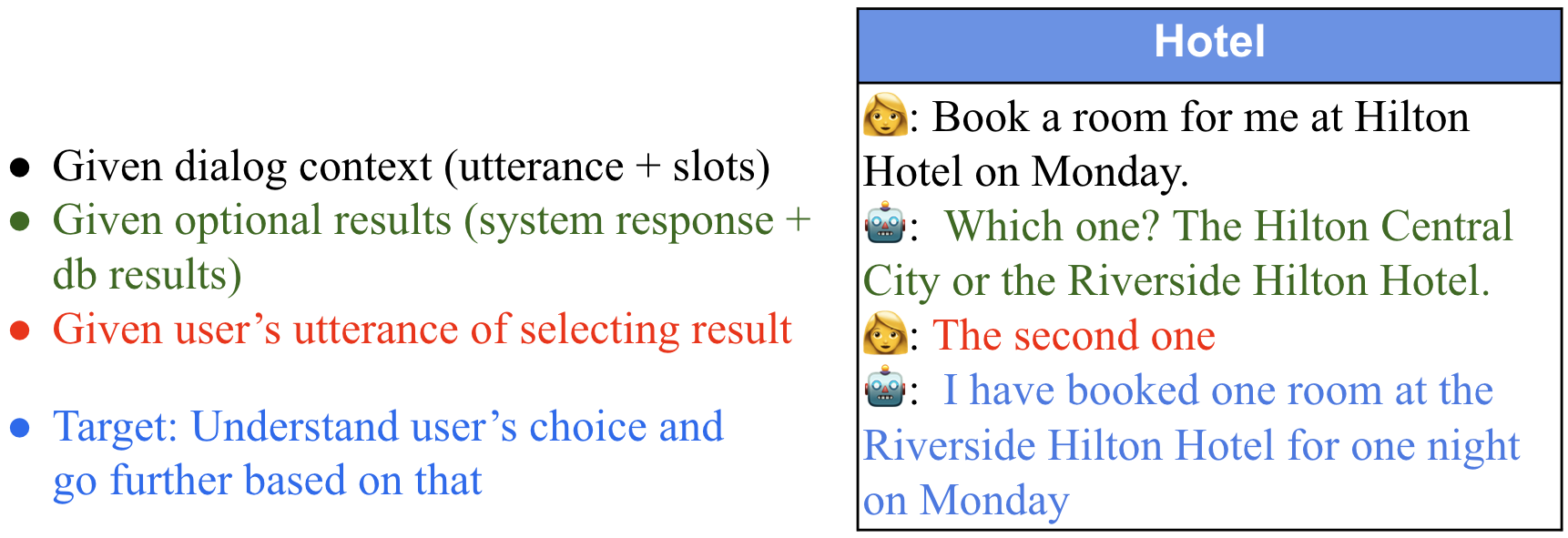}
\caption{For this disambiguation task, we assume the dialog context, system utterance including result list and user's answer are given. The goal is to extract the result that the user select and continue the dialog.
}
\label{fig_task_form}
\end{figure}

In this paper, we propose a new task called disambiguation in dialog database search. 
As shown in Fig.~\ref{fig_task_form}, the task assumes that we are provided with the dialog context $c$, the system response $s$ which includes all the optional results 
, and the user's utterance $u$ that make a choice. To avoid redundant option lists, we limit the number of options to less than five. The target of the task is to extract the entity of the result selected by the user.

\section{Dataset}
The most popular task-oriented dialog datasets (MultiWOZ, SGD) do not contain many cases for the disambiguation task. 
In order to enable the dialog model to handle this task, we propose to augment these two datasets in three steps described in the following subsections.
\subsection{Synthesizing Single-Turn Dialog}
\label{sec:syn_data}

\begin{figure}[t]
\centering
\includegraphics[width=0.97\columnwidth]{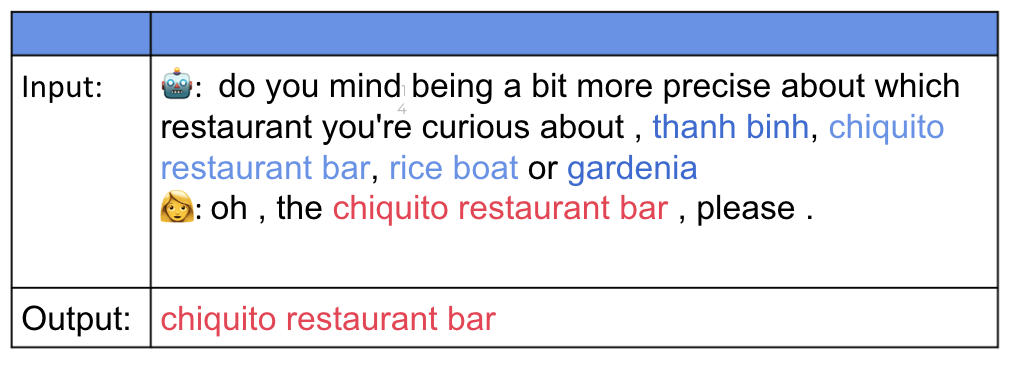}
\caption{An example of the synthesized single-turn dialog. The utterance templates are generated based on CFGs. 
The \textit{candidate entities} (italicized) are sampled from the database of MultiWOZ or SGD. The \textbf{selected entity} (bolded) is sampled from the candidates.
}
\label{fig_single_turn}
\end{figure}

We first develop a single-turn dialog dataset. 
With this single-turn dataset, the fine-tuned dialog model can focus only on the disambiguation turns and learn the skill to solve the ambiguity problem.
Fig.~\ref{fig_single_turn} shows an example of the dialog turn, which we would use through this section to introduce the dataset.
In this dataset, each dialog turn consists of only a system utterance and a user response. The system utterance gives a list of options (marked in blue) and the user response makes a choice from the list (marked in red). The ground truth output is the name entity of the chosen result. 

To synthesize the system and user sentences, we extracted templates from disambiguation turns from the SIMMC 2.0 dataset. For example, the system from SIMMC2.0 asks questions like \textit{``do you mind being a bit more precise about which shoes you're curious about, the red one or the blue one''} to solve ambiguity. We delexicalize those utterance by removing the all domain-related tokens such as \textit{``shoes'', ``the red one'', ``the blue one''}, and keep the rest as a template.
We then extract a list of context-free grammars (CFGs) from those templates, and then generate natural sentences based on the CFGs.
For example, from the previous template we can summarize a grammar:\textit{``SENT -> do you mind VERBING''}, where \textit{``VERBING''} is a non-terminal token for a verb phrase in an ``ING'' form. 
The CFG-based generator can potentially generate around 2 million different system questions and 30K+ different user utterances, which ensure the diversity of the generated data. 
To cover multiple domains, we utilize the database from the MultiWOZ and SGD datasets, which in total covers 27 domains, each containing one name entity type. 
We randomly sample a certain number of values from the database based on the domain and entity type, and insert them into the system response. 
The number of candidate values is also randomly sampled. 
To make the sentence more natural, we limit the candidate number to be between three and five. 
Then, we randomly sample one from the candidate list as the selected result.

To make the task harder and more realistic, we also explore different entity addressing methods to generate the user utterance:
\begin{itemize}
    \item \textbf{Positional Addressing.} 
    Instead of directly addressing the name entity (Fig.~\ref{fig_single_turn}), users use entity's list position, e.g., ``the second one''.
    \item \textbf{Partial Addressing.} 
    User use part of the name for simplicity, e.g. ``chiquito'' instead of ``chiquito restauraant bar''
    \item \textbf{Addressing with Typo.} 
    We add typos in the name entity to make the model more robust.
    \item \textbf{Multiple Addressing.} 
    User chooses more than one option at a single time and the model is expected to extract all their choices.
    \item \textbf{Addressing with Attributes.} 
    User describes the selected result with more attributes, e.g. ``the restaurant in the north of the city''.
\end{itemize}

\begin{table*}[t]
\centering
\small
\begin{tabular}[width=\textwidth]{l|ccc|ccc}
\toprule \hline
      & \multicolumn{3}{c|}{SGD}      & \multicolumn{3}{c}{MultiWOZ}        \\ \cline{2-7} 
      & train       & dev       & test       & train       & dev       & test      \\  \hline
dialog & 4.7k / 16k  & 0.9k / 2.5k  & 1.6k / 4.2k  & 2.7k / 8.4k   & 0.3k / 1k  & 0.3k /1k  \\ 
turn   & 5.1k / 330k & 1.0k / 48.7k & 1.8k / 84.6k & 3.2k / 105k & 0.4k / 13.8k & 0.4k / 13.7k \\  \hline
\bottomrule
\end{tabular}
\caption{Statistics for disambiguation augmentation. The table presents the numbers of dialogs or turns that are modified for disambiguation cases, and the numbers on the right side of slash are the total number of dialogs or turns in each dataset. 
}
\label{table:stats_aug}
\end{table*}

\subsection{Automatic Augmentation}
\label{sec:auto_aug}
The single-turn dialog dataset helps enable models to solve the disambiguation task.
However, the single-turn is not an entire dialog and the model barely trained with that can hardly conduct a complete dialog. Our goal is to enhance a complete dialog model with the disambiguation skill while keeping the performance of other tasks. 
Currently, most of the state-of-the-art task-oriented dialog models are trained with MultiWOZ and SGD dataset. 
Therefore, we propose to augment these two dataset by adding disambiguation turns.

\begin{figure}[t]
\centering
\includegraphics[width=0.97\columnwidth]{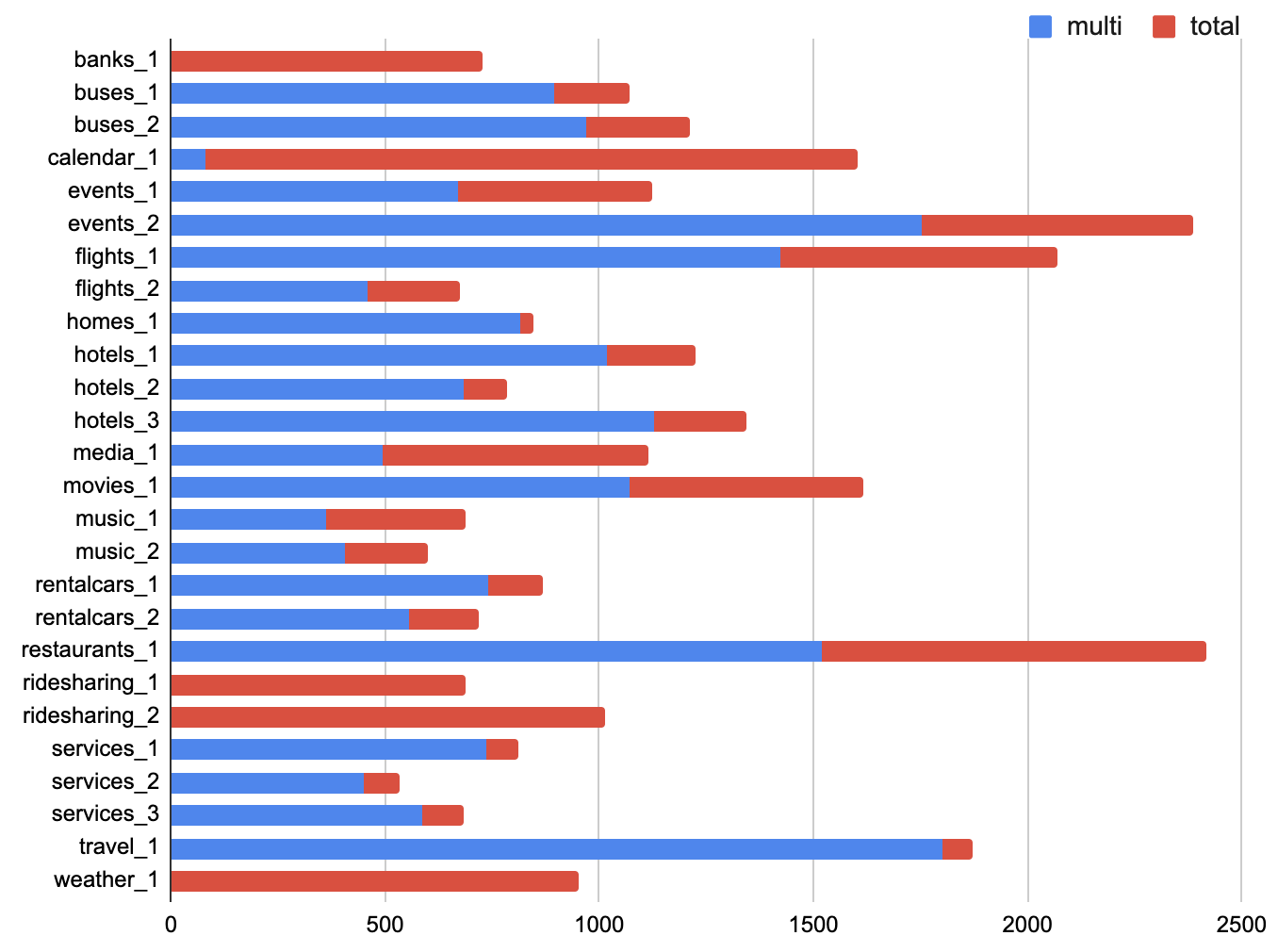}
\caption{The blue bar represents the number of dialogs which contain multiple database-search results in each service from the SGD dataset. While the red bar represents the total number of dialogs in each service.
}
\label{fig_sgd}
\end{figure}

Fig.~\ref{fig_sgd} shows the proportion of the dialogs in each domain that contains multiple results.
We find that nearly 66.7\% of dialogs involve multiple results, where ambiguity can occur.
Though in both SGD and MultiWOZ, system would always give a suggestion after searching the database, e.g. ``I have 10 suitable results, how about ...'' and the user side would simply accept it or ask about something else. 
This avoids the ambiguity in the dataset. 
However, the system in the reality would still face the ambiguity problem when interacting with real human beings, who would like to know more about other options. 
Therefore, we want to augment these two popular dataset with disambiguation turns to improve the model's ability.

\begin{figure}[t]
\centering
\includegraphics[width=\columnwidth]{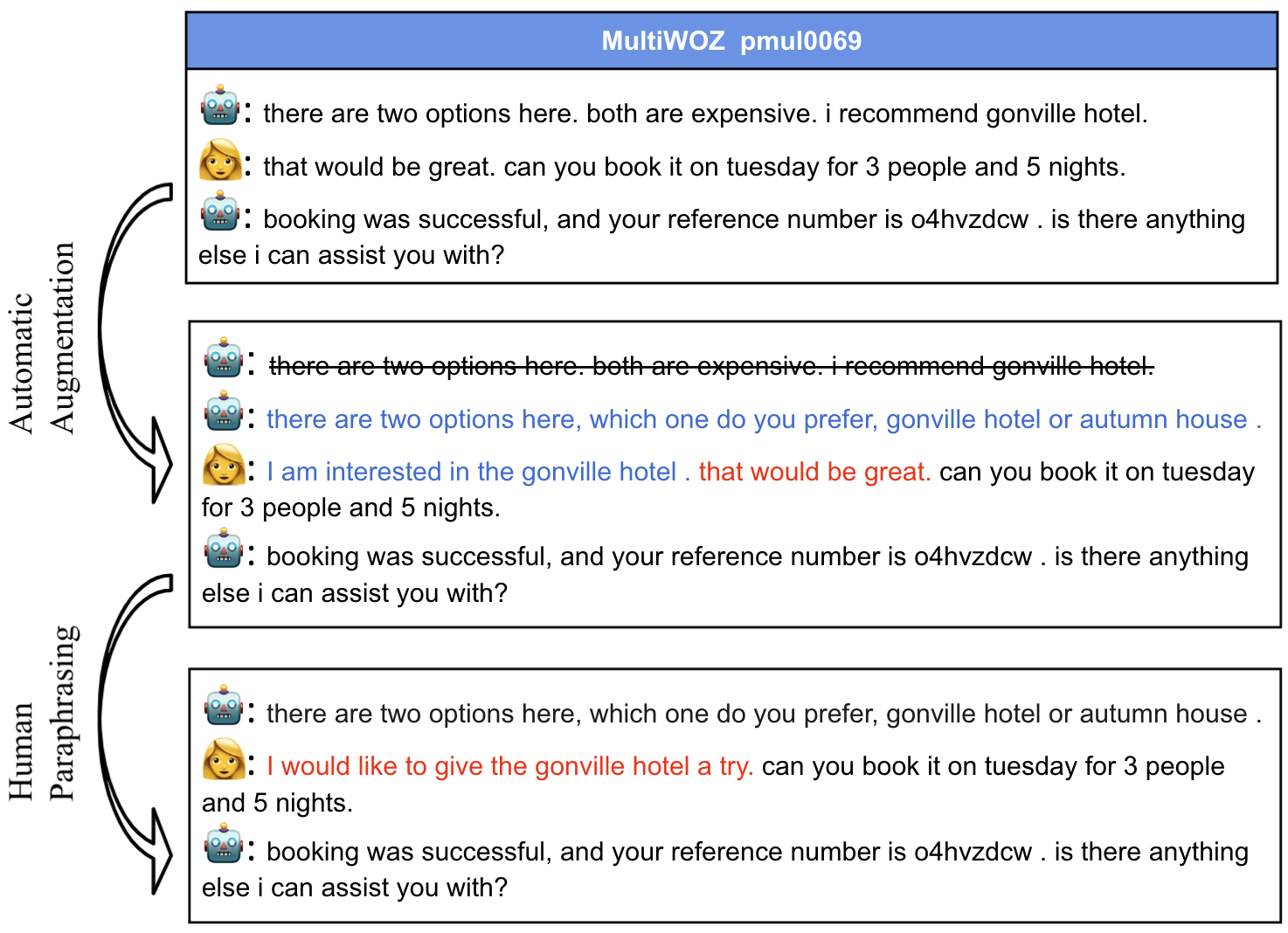}
\caption{An example of the automatic disambiguation augmentation and human paraphrasing. We first replace the original system suggestion with a synthesized utterance, listing all candidate entities and asking user to select. Then, we generate user chosen answer and insert it to the beginning of the original user utterance. For human paraphrasing, we ask crowd-workers to rewrite the user utterance to gain naturalness and diversity.
}
\label{fig_aug_eg}
\end{figure}

First, we locate the turns to be modified. In those turns, the system presents the database-searching results, where the ambiguity takes place. 
We also incorporate relevant annotation and sentence structure to filter out some inappropriate cases, e.g. the user does not make any choices in this turn.
Then we generate a new system utterance to replace the original one. The generation is conducted based on the same toolkit and CFGs from Sec.~\ref{sec:syn_data}, and the slot values are extracted from the corresponding database. 
As shown in Fig.~\ref{fig_aug_eg} (highlighted in blue), the new system utterance provides a list of specific searching results without giving any suggestion. 
Following the system utterance, a user utterance is also generated to make the choice, which should be consistent with the original suggestion that the user accepts. 
If the user rejects the original system suggestion, we do not make any modification. In the end, we concatenate the generated user utterance with the original one. 
In this way, we ensure the other unchanged turns of the dialog (especially the following turns) will be coherent with the modified turns, in order to eliminate the effects on the unchanged turns of the dialog as much as possible.

We conduct the same progress on both SGD and MultiWOZ dataset. 
Note that the ambiguity problem occurs only when there is a specific target entity, e.g. hotel name in the ``hotel'' domain and not
every domain includes such an entity (e.g. any car satisfying constraints is acceptable in the ``taxi'' domain). 
Therefore, we only augment the ``restaurant'', ``hotel'', and ``attraction'' domains in the MultiWOZ dataset, and 24 out of 45 services in the SGD dataset, which are listed in the Appendix~\ref{sgd_services}.
The statistics of the augmentation is listed in the Table.~\ref{table:stats_aug}. 
More than 30\% of dialogs are involved and with disambiguation turns, and around 2\% of the turns are modified.

The newly generated user utterance is simply the concatenation of the template utterance and the original utterance that responds to the system suggestion. 
Therefore, the connection between them can be unnatural. 
In addition, the new user utterance is generated by CFG, which means the utterance itself can be unnatural. Therefore, we conduct human paraphrasing to improve the quality of the user utterance.

\subsection{Human Paraphrasing}
\label{sec:human_aug}
We recruit crowd-workers to paraphrase the disambiguation turns. 
Before starting the paraphrasing job, each crowd-worker is required to read through a guideline document to get a better understanding of the task, the requirements and the workflow. 
A screenshot of the paraphrasing interface is shown in the Appendix Fig.~\ref{fig_human}. 
For each paraphrasing job, we present a good example of paraphrasing in the same page as the turn to be modified.
To keep consistent with task description in the Sec.~\ref{sec:task_form}, we provide the crowd-workers with 1) the modified system utterance, which includes a list of options and asks the user to select, 2) the user utterance, which concatenates the template-generated sentence and the original user utterance. 
In the interface, the user utterance is highlighted in a different color (green) and marked as ``need paraphrase''. 
To avoid changing user's original choice during paraphrasing, we also show crowd-worker the result value that the user should choose, keeping consistent with the dialog state annotation. 
In addition, to ensure the disambiguation turn is coherent with the dialog context, we also present the previous user utterance and the next system response.

We conduct the paraphrasing job for the test sets from both SGD and MultiWOZ, as well as the training set of SGD.
To evaluate the quality of the human paraphrase process, we randomly sample 5\% of the disambiguation turns and ask another group of crowd-workers to judge whether the modification is valid, which means satisfying all the requirements listed in the guideline document (maintaining all essential information, not similar to the original utterance, not natural, etc.). 
Each turn receives two judgements. 
In total, we have an 88\% of agreement rate between two judgements and 92\% of the agreements are error free, which means our paraphrasing job is valid.
We also ask annotators to point out if there is any ethical violation in the utterance, which is discussed in more details in Sec.~\ref{sec:ethic}.

\begin{table*}[t]
\centering
\small
\begin{tabular}[width=\textwidth]{l|ccc|ccc}
\toprule \hline
\multirow{2}{*}{\diagbox{Train Data}{Test Data}} & \multicolumn{3}{c}{SGD  (only aug turns)} & \multicolumn{3}{|c}{MultiWOZ  (only aug turns)} \\ \cline{2-7} 
                  & Origin    & AutoAug    & HumanAug & Origin    & AutoAug    & HumanAug    \\
\hline
Origin            & 0.556$_{\pm 0.007}$          & 0.242$_{\pm 0.006}$          & 0.211$_{\pm 0.008}$          & \textbf{0.676$_{\pm 0.007}$} & 0.488$_{\pm 0.005}$          & 0.488$_{\pm 0.001}$           \\
Origin+Syn2\%     & \textbf{0.575$_{\pm 0.014}$} & 0.279$_{\pm 0.025}$          & 0.252$_{\pm 0.018}$          & 0.650$_{\pm 0.003}$          & 0.489$_{\pm 0.014}$          & 0.494$_{\pm 0.016}$           \\
Origin+Syn100\%   & 0.571$_{\pm 0.006}$          & 0.344$_{\pm 0.018}$          & 0.304$_{\pm 0.015}$          & 0.670$_{\pm 0.007}$          & 0.554$_{\pm 0.026}$          & 0.556$_{\pm 0.029}$           \\
AutoAug           & 0.551$_{\pm 0.002}$          & 0.496$_{\pm 0.005}$          & 0.437$_{\pm 0.008}$          & 0.633$_{\pm 0.016}$          & 0.744$_{\pm 0.025}$          & 0.739$_{\pm 0.029}$           \\
AutoAug+Syn2\%    & 0.569$_{\pm 0.004}$          & 0.548$_{\pm 0.010}$          & 0.488$_{\pm 0.017}$          & 0.642$_{\pm 0.007}$          & 0.838$_{\pm 0.002}$          & 0.830$_{\pm 0.007}$           \\
AutoAug+Syn100\%  & 0.567$_{\pm 0.009}$          & \textbf{0.583$_{\pm 0.001}$} & \textbf{0.501$_{\pm 0.002}$} & 0.633$_{\pm 0.013}$          & \textbf{0.846$_{\pm 0.002}$} & \textbf{0.837$_{\pm 0.007}$}  \\
\hline \bottomrule
\end{tabular}
\caption{The accuracy of the name entity prediction for only the augmented turns. Each number represents the performance of a model trained with a certain training data setting and evaluated on a certain test set. ``Origin''/``AutoAug''/``HumanAug'' represents evaluation on the original/automatic augmented(Sec.~\ref{sec:auto_aug})/human paraphrased(Sec.~\ref{sec:human_aug}) data. ``+Syn'' represents mixed with synthesized data and the percentage following ``+Syn'' means the amount of synthesized data compare to the whole test set.
}
\label{table_main}
\end{table*}

\section{Experiment}
We use GPT2~\cite{Radford2019LanguageMA} as our backbone model and fine-tune it with the augmented SGD and MultiWOZ datasets separately.

\paragraph{MultiWOZ.} MultiWOZ~\cite{budzianowski-etal-2018-multiwoz} is a multi-task task-oriented dialog dataset.
It covers seven domains and contains 10K+ dialogs. Our augmentation focuses mainly on three domains:``\textit{attraction}'', ``\textit{hotel}'' and ``\textit{restaurant}'', involving more than 3K dialogs. 
We choose to conduct our augmentation based on the MultiWOZ 2.2~\cite{zang-etal-2020-multiwoz}, which is the most widely-accepted version.

\paragraph{Schema-Guided Dataset.} SGD~\cite{rastogi2019towards} is another popular multi-task dialog dataset.
Since the DSR-ambiguity problem requires the service containing a target entity and not every service satisfies that requirement, our augmentation involved totally 10 domains and 24 services. 

We directly compute the accuracy on whether the model can successfully predict the correct name entity as evaluation metric. 
Since the generation is similar to the dialog state tracking task, we also compute the joint goal accuracy (details in Appendix.\ref{jga}) to evaluate whether the augmentation maintain the model's performance of other tasks.

We train GPT2 with both the original and augmented data, and test the fine-tuned models on original/augmented/human paraphrased test sets.
The same experiment is conducted for both datasets.
In addition to original and augmented training data, we also explore the impact of the synthesized single-turn dialog. 
Learned from Table~\ref{table:stats_aug}, the augmented turns only take up 2\% of the whole dataset.
In order to achieve a similar amount of augmentation compared to the automatic augmented data, we sample $5k$ synthesized single-turn dialogs for SGD and $3k$ for MultiWOZ, which is around 2\% of each training set.
Then, we mix those dialogs with the original (or augmented) training data and evaluate on three test data settings.
We also increase the sampling amount of the synthesized dialog to be comparable to the whole training set, represented by ``Syn100\%'' in the table, to explore whether the model achieves a better learning of the entity disambiguation skill with access to more disambiguation cases.

\section{Results and Analysis}
In this section, we present our experimental results including key observations and ablation studies. In addition, we also analyze how to leverage our augmented dataset to deal with DSR-ambiguity in new datasets.

\subsection{Augmentation Helps Resolve Ambiguity}
Table~\ref{table_main} shows the name entity prediction accuracy evaluated only on the turns involved in augmentation, which is around 2\% of the whole test set.
The first column states the different training data settings that we use to fine-tune the GPT2 model, and the first row presents three different test sets.

Comparing the ``Origin'' column and ``AutoAug'' column, we find that the performance of the model trained with original data drastically drops from 0.556 to 0.242 for SGD and from 0.676 to 0.488 for MultiWOZ. 
This verifies our hypothesis that the original datasets contain few disambiguation cases.
Therefore, the model trained with the original data cannot understand user's answer towards the clarification question and extract the corresponding entity tokens.
On the other hand, the models trained with augmented data achieve better performance (from 0.242 to 0.496 for SGD and from 0.488 to 0.744 for MultiWOZ) on the augmented data, which means those models learn the skill to complete the disambiguation task.
The results on the human paraphrased test set, which is more diverse and natural, support the same conclusion.
We also combine the synthesized single-turn dialog data with the original training data (or the augmented training data). 
The original data mixed with full-size synthesized data setting achieves the best result on human paraphrased test set for SGD and the augmented data mixed with full-size synthesized data setting achieves the best one for MultiWOZ. 

Table~\ref{table_name_full} shows the overall name entity accuracy of the whole test set. 
Since the augmentation only modifies 2\% turns of the whole test set, the difference between the performance of on the original and augmented test set is not as apparent as Table~\ref{table_main}. 
However, the model trained with augmented data still performs better than the model trained with original data on both augmented and human paraphrased test set. 
The model under ``Aug+Syn100\%'' train setting achieves the best results on five out of six test sets, showing that the augmentation and synthesized data jointly enhance the model's ability to extract name entity.
In addition to name entity prediction, we also explore whether the augmentation helps the model to predict other slot types by computing the joint goal accuracy.
Table~\ref{table_dst_aug} shows the results for only the augmented turns and Table~\ref{table_dst_full} lists the results on the whole test set. 
In both tables, the setting ``Aug+Syn100\%'' achieves the best or the second best performance for both augmented and human paraphrased test sets.
Hence, our augmentation not only enables the model to solve the disambiguation task, but also improves its ability for dialog state tracking task.
The improvement mainly results from the similarity of the disambiguation task and the dialog state tracking, and more augmented data points enhance the model's understanding of the input sequence.

\begin{table*}[h]
\centering
\small
\begin{tabular}[width=\textwidth]{l|ccc|ccc}
\toprule \hline
\multirow{2}{*}{\diagbox{Train Data}{Test Data}} & \multicolumn{3}{c}{SGD} & \multicolumn{3}{|c}{MultiWOZ} \\ \cline{2-7} 
                  & Origin    & AutoAug    & HumanAug & Origin    & AutoAug    & HumanAug    \\ \hline
Origin          & 0.489$_{\pm 0.007}$          & 0.477$_{\pm 0.007}$          & 0.477$_{\pm 0.007}$          & \textbf{0.535$_{\pm 0.001}$} & 0.522$_{\pm 0.005}$          & 0.523$_{\pm 0.004}$          \\
Origin+Syn2\%   & 0.500$_{\pm 0.003}$          & 0.489$_{\pm 0.004}$          & 0.490$_{\pm 0.004}$          & 0.530$_{\pm 0.001}$          & 0.500$_{\pm 0.006}$          & 0.501$_{\pm 0.006}$          \\
Origin+Syn100\% & 0.495$_{\pm 0.006}$          & 0.487$_{\pm 0.005}$          & 0.487$_{\pm 0.005}$          & 0.528$_{\pm 0.003}$          & 0.504$_{\pm 0.005}$          & 0.504$_{\pm 0.004}$          \\
AutoAug             & 0.502$_{\pm 0.010}$          & 0.499$_{\pm 0.010}$          & 0.497$_{\pm 0.010}$          & 0.524$_{\pm 0.004}$          & 0.535$_{\pm 0.003}$          & 0.535$_{\pm 0.003}$          \\
AutoAug+Syn2\%      & 0.498$_{\pm 0.004}$          & 0.496$_{\pm 0.004}$          & 0.494$_{\pm 0.004}$          & 0.525$_{\pm 0.002}$          & 0.545$_{\pm 0.001}$          & 0.545$_{\pm 0.001}$          \\
AutoAug+Syn100\%    & \textbf{0.510$_{\pm 0.004}$} & \textbf{0.509$_{\pm 0.004}$} & \textbf{0.506$_{\pm 0.004}$} & 0.532$_{\pm 0.002}$          & \textbf{0.552$_{\pm 0.004}$} & \textbf{0.552$_{\pm 0.004}$}\\ \hline \bottomrule
\end{tabular}
\caption{Joint goal accuracy evaluated on the whole test set.}
\label{table_dst_full}
\end{table*}

\begin{table*}[t]
\centering
\small
\begin{tabular}[width=\textwidth]{l|ccc|ccc}
\toprule \hline
\multirow{2}{*}{\diagbox{Train Data}{Metrics}} & \multicolumn{3}{c}{JGA over Whole Test Set} & \multicolumn{3}{|c}{Name Entity Accuracy for Aug Turns} \\ \cline{2-7} 
                  & Origin    & AutoAug    & HumanAug & Origin    & AutoAug    & HumanAug    \\
\hline
Origin       & 0.535$_{\pm 0.001}$ & 0.522$_{\pm 0.005}$ & 0.523$_{\pm 0.004}$ & 0.676$_{\pm 0.007}$ & 0.488$_{\pm 0.005}$ & 0.488$_{\pm 0.001}$ \\
Origin+Syn      & 0.528$_{\pm 0.003}$ & 0.504$_{\pm 0.005}$ & 0.504$_{\pm 0.004}$ & 0.670$_{\pm 0.007}$ & 0.554$_{\pm 0.026}$ & 0.556$_{\pm 0.029}$ \\
Aug          & 0.524$_{\pm 0.004}$ & 0.535$_{\pm 0.003}$ & 0.535$_{\pm 0.003}$ & 0.633$_{\pm 0.016}$ & 0.744$_{\pm 0.025}$ & 0.739$_{\pm 0.029}$ \\
Aug+Syn      & 0.532$_{\pm 0.002}$ & \textbf{0.552$_{\pm 0.004}$} & \textbf{0.552$_{\pm 0.004}$} & 0.633$_{\pm 0.013}$ & 0.878$_{\pm 0.000}$ & 0.874$_{\pm 0.004}$ \\
SGDori+Origin       & 0.537$_{\pm 0.002}$ & 0.517$_{\pm 0.001}$ & 0.518$_{\pm 0.002}$ & 0.678$_{\pm 0.004}$ & 0.441$_{\pm 0.012}$ & 0.441$_{\pm 0.013}$ \\
SGDaug+Origin       & 0.534$_{\pm 0.006}$ & 0.522$_{\pm 0.006}$ & 0.522$_{\pm 0.007}$ & 0.684$_{\pm 0.003}$ & 0.505$_{\pm 0.012}$ & 0.495$_{\pm 0.011}$ \\
SGDaug+Origin+Syn   & \textbf{0.540$_{\pm 0.003}$} & 0.529$_{\pm 0.005}$ & 0.529$_{\pm 0.005}$ & \textbf{0.685$_{\pm 0.009}$} & 0.650$_{\pm 0.009}$ & 0.658$_{\pm 0.006}$ \\
Upsample     & 0.528$_{\pm 0.002}$ & 0.544$_{\pm 0.002}$ & 0.543$_{\pm 0.002}$ & 0.635$_{\pm 0.010}$ & 0.846$_{\pm 0.030}$ & 0.837$_{\pm 0.032}$ \\
Upsample+Syn & 0.526$_{\pm 0.002}$ & 0.543$_{\pm 0.001}$ & 0.542$_{\pm 0.001}$ & 0.633$_{\pm 0.005}$ & \textbf{0.884$_{\pm 0.007}$} & \textbf{0.883$_{\pm 0.008}$} \\
\hline \bottomrule
\end{tabular}
\caption{Results for more training setting based on the MultiWOZ dataset, where left part is the joint goal accuracy of the whole test set and right part is name entity accuracy over only augmented turns. The amount of synthesized data ``+Syn'' is the same as the amount of original test test in this table. ``SGDori'' means first fine-tuning model with original SGD training data and then fine-tuning on MultiWOZ training data, while ``SGDaug'' means first fin-tuning model with augmented SGD training data. The setting ``Upsample'' means up-sampling augmented turns to the same amount of training data.
}
\label{table_multiwoz}
\end{table*}

\subsection{Augmentation Brings No Harm}
Our ultimate goal is to expand end-to-end task oriented dialog systems with the disambiguation skill. 
Therefore, it is required not only to enable the dialog model to resolve DSR-ambiguity, but also to maintain the model's original ability for generating responses or dialog state tracking.
To verify that, we first analyze the performance on the original test set (``Origin'' columns in Table~\ref{table_main}).
The models trained with original data (0.676 on MultiWOZ) or the original one mixed with 5\% synthesized data (0.575 on SGD) commonly achieves the best performance, which is reasonable since training data and test data share almost the same distribution. 
On the other hand, the performance on the original test set of the models trained with the augmented data is comparable with the original training data, which means these models maintain the ability to predict entity name. 
As for the results over the whole test set in Table~\ref{table_name_full}, the augmented model even achieves better accuracy (0.877) than the original one (0.871) on the SGD test set.
Therefore, the augmentation does not hurt the model's ability to predict name entities without disambiguation cases.

Beyond name entities, the augmentation hardly affects the model's ability to predict other dialog slots for the non-disambiguation cases. 
The results are listed in the ``Origin'' columns in the Table~\ref{table_dst_aug} and Table~\ref{table_dst_full} correspondingly. 
For both test sets, the models trained with augmented data achieve comparable results with the models trained with original data, which means our augmentation also maintains the distribution of other slot types in the original data.
In conclusion, our augmentation does not impede the model from learning the original data distribution.
And the model trained with the augmented data perform well no matter whether the disambiguation case exists.

\subsection{Leveraging Augmented Turns}
To find the most efficient method to leverage our dataset, we explore the following experiment settings. Since SGD and MultiWOZ are both task-oriented dialog datasets and share some common domains, pre-training on one dataset might help learn the other one. Therefore, we first fine-tune the model with the original SGD and then fine-tune it on the MultiWOZ. We also conduct the experiment that uses the augmented SGD training data for the first step of fine-tuning. Since the augmented turns only take up 2\% of the whole training data, the model rarely sees the disambiguation cases in each epoch. To emphasize those turns, we up-sample those disambiguation turns to the same amount as the original training data. 

Table~\ref{table_multiwoz} shows results for these settings. For the name entity accuracy, the setting ``Upsample+Syn'' achieves the best result, because the more disambiguation turns the models see, the better the model learns the skill to solve the ambiguity. As for the joint goal accuracy, setting ``Aug+Syn'' performs better than ``Upsample+Syn'' because too much disambiguation turns inevitably introduce bias and affect learning the original task. Therefore, if we need to solve DSR-ambiguity in a new dataset, the best option is to conduct augmentation with our framework and train models together with synthesized single-turn data. Although not as good as setting ``Aug+Syn'', the setting ``SGDaug+Origin+Syn'' performs better than the model trained on original data in terms of both JGA and name entity accuracy. Notice that this setting does not require any augmentation on MultiWOZ. Hence, to solve disambiguation cases in a new dataset, the cheapest choice is to fine-tune a model on our augmented dataset (MultiWOZ and SGD) first, and then fine-tune it on the original data, mixed with the synthesized single-turn dataset.

\subsection{Impact of Entity Addressing Methods}
\label{sec:result_single_turn}
To explore the impact of different addressing methods, we conduct the ablation study by fine-tuning GPT2 with the synthesized single-turn dialog datasets of each individual addressing method.
For each addressing method, we generate $100K$/$10K$/$10K$ single-turn dialogs as the train/dev/test set, which is comparable to the MultiWOZ or the SGD datasets. 
Besides each single addressing method, we also explore different combinations of the addressing methods.
We fine that when focusing only on the disambiguation task with a simple context structure like single-turn dialog, the model can easily learn all kinds of addressing methods, except for ``multiple addressing''. The model accuracy drops by $\approx 33\%$ in that case.
Even if we combine multiple addressing methods together except ``multiple addressing'', the model can still understand the addressing target. 
However, when the user chose multiple entities, it is hard for models to accurately predict how many entities the user selected.

\section{Related Work}
\label{sec:related_work}

\subsection{Task-Oriented Dialog Datasets}
MultiWOZ~\cite{budzianowski-etal-2018-multiwoz} is one of the most popular task-oriented dialog dataset. It covers multiple domains, consists of a large amount of dialogs, and has been chosen as benchmark for many dialog tasks, e.g. dialog state tracking~\cite{Zhang2019FindOC, Zhang2020APE, Heck2020TripPyAT}, dialog policy optimization~\cite{Wu2019AlternatingRD, Wang2020ModellingHS, Wang2020MultiDomainDA} and end-to-end dialog modeling~\cite{Zhang2020TaskOrientedDS, HosseiniAsl2020ASL, Peng2020SOLOISTFT, huang2021dair}. And to polish it up to be a better benchmark, many works pay effort to improve and correct dataset~\cite{eric-etal-2020-multiwoz, zang-etal-2020-multiwoz, qian-etal-2021-annotation, Han2021MultiWOZ2A, Ye2021MultiWOZ2A}. In this paper, we choose MultiWOZ 2.2 version to conduct augmentation. 

Schema-Guided Dataset (SGD)~\cite{rastogi2019towards} is the largest public task-oriented dialog dataset, containing 18K+ dialogs. It covers 20 domains and each domain consists of multiple services, e.g. ``hotel'' domain can either ask for hotel's information or make a reservation. And in total, there are 45 services. The dataset is constructed by generating dialog outlines from interactions between two dialog simulators, and then being paraphrased by crowd-workers.

SIMMC 2.0~\cite{Kottur2021SIMMC2A} is a newly-released multi-modal task-oriented dialog dataset around situated interactive multi-modal conversations~\cite{moon2020situated}. It focuses on dialogs with multi-modal context, which can be in the form of either co-observed image or virtual reality environment. The dataset contains 11K+ dialogs and covers two shopping domains. 

As for the disambiguation problem, neither MultiWOZ nor SGD has related cases or annotations. SIMMC 2.0 is well-annotated for disambiguation, but it only covers two domains, and addresses entity mostly with multi-modal knowledge. Therefore, we augment MultiWOZ and SGD with the disambiguation templates from the SIMMC 2.0.

\subsection{Ambiguity \& Clarification Questions}

Ambiguity is a common phenomenon across many conversation-involved NLP tasks, e.g. conversational search~\cite{Rosset2020LeadingCS}, Question-Answering~\cite{white-etal-2021-open}, open-domain dialog~\cite{aliannejadi-etal-2021-building} and intent classification~\cite{Bihani2021FuzzyCO, Dhole2020ResolvingIA}.
The problem mainly results from two aspects: 1. user's ambiguous keyword (e.g. ``orange'' can be either color or fruit~\cite{Coden2015DidYM}) and 2. lacking of enough constraints for accurate searching, leading to multiple results (e.g.``I want to book a cheap hotel'' where there might be multiple ``cheap'' hotels).
Previous work proposes to incorporate clarification questions to solve the ambiguity problem~\cite{purver-etal-2001-means, schlangen-2004-causes, Radlinski2017ATF}, including both model-wise~\cite{Li2017LearningTD, rao-daume-iii-2019-answer,yu-etal-2020-interactive} and dataset-wise~\cite{Aliannejadi2019AskingCQ, xu-etal-2019-asking, min-etal-2020-ambigqa, Zamani2020MIMICSAL}. 
Our work it the first to point out the ambiguity within the database-searching of task-oriented dialog systems and introduce clarification questions to help solve this problem. 

In addition, most of the work focus on when and how to generate clarification questions~\cite{kumar-black-2020-clarq}.
Typical clarification question generation is based on the context with a Seq2Seq model~\cite{Zamani2020GeneratingCQ}.
\citet{rao-daume-iii-2019-answer} propose to utilize the generative adversarial network to learn generating relevant clarification question based on corresponding answers.
\citet{Sekulic2021UserEP} takes user engagement into consideration to generate high-quality clarification questions.
In this work, instead of focusing on question generation, we put our attention on understanding the user's answer to clarification questions.

\section{Conclusion and Future Work}
In this paper, we propose a realistic task, disambiguation, which is ignored in most popular public task-oriented dialog datasets such as MultiWOZ and SGD. Therefore, models trained based on these two datasets can not deal with entity ambiguity. We propose augmenting these two datasets to create datasets address the ambiguity problem. We extract templates of the disambiguation turns from the SIMMC 2.0 dataset and jointly generate new turns with the databases from MultiWOZ and SGD for augmentation. To ensure the quality and correctness of the augmentation, we recruit crowd-workers to paraphrase the generated sentences. We benchmark our augmented dataset with the GPT2 model. The result suggests that the augmented dataset not only strengthens dialog models with a new skill to solve disambiguation tasks, but also does not impact the performance on the original tasks.

However, the current performance of the prediction still leaves space for improvement and the augmentation for the disambiguation task does not cover all possible cases in practice. Therefore, in the future, we plan to incorporate state-of-the-art entity reference techniques to improve the model's performance. In addition, we would explore more realistic disambiguation cases to improve the datasets, which further enhances the dialog system. By incorporating more and more realistic cases into the training data, we hope this work will help to build dialog systems that handle more complex tasks in daily lives.

\section{Ethical Considerations}
\label{sec:ethic}
To ensure that the dataset does not have any sensitive topics, we ask crowd-workers to make comments if the dialog content involves any of following: 1. offensive, racist, biased and non-tolerant behavior; 2. violence and self-harm; 3. sexual or flirtatious behavior; 4. controversial and polarizing topics. Since the database of both MultiWOZ and SGD are sampled from real world, annotators also comment if there are real names included in the slot values, which can be personally identifiable information (PII). Considering both of these two datasets are public dataset, we do not replace those name entities with placeholders. The detailed description of sensitive topics is included in the Fig.~\ref{fig_ethical} in the appendix.

\bibliography{anthology,custom}
\bibliographystyle{acl_natbib}

\newpage
~\newpage
\onecolumn
\appendix

\section{Supplementary Details for Augmentation}
\subsection{Involving Domains}
\begin{itemize}
    \item \textbf{MultiWOZ: } ``restaurant'', ``hotel'', and ``attraction''
    \item \textbf{Google SGD: } 
    ''events\_3'',
    ''homes\_2'',
    ''hotels\_4'',
    ''media\_3'' ,
    ''messaging\_1'' ,
    ''movies\_1'',
    ''movies\_3'',
    ''music\_3'',
    ''restaurants\_2'',
    ''services\_1'',
    ''services\_4'',
    ''travel\_1'',
    ''events\_1'',
    ''homes\_1'',
    ''hotels\_1'',
    ''media\_2'',
    ''movies\_2'',
    ''music\_1'',
    ''hotels\_3'',
    ''media\_1'',
    ''music\_2'',
    ''restaurants\_1'',
    ''services\_2'',
    ''services\_3'',
\label{sgd_services}
\end{itemize}

\subsection{Human Paraphrasing}
The whole paraphrasing job involved 37 annotators and cost around \$26,000 in total.

\newpage
\section{Supplementary Details for Experiments}

\subsection{Hyper-Parameters}
We do a hyper-parameter search for the training on both original dataset and augmented dataset and find the following setting: a batch size of 4 and learning rate of 5e-6 is the best one for both. 
We run at most 20 epochs for each experiment and do validation for every epoch, with an early stop step of 3. %
For each experiment, we run for three times with different random seeds and report the average value, along with the standard deviation.

\subsection{Metric}
\label{jga}
\noindent\textbf{Joint Goal Accuracy} evaluates the performance of predicting dialog states. It counts one for each turn if the model successfully generate all slot values, otherwise count zero.
\subsection{Supplementary Experiment Results}
\begin{table*}[h]
\centering
\small
\begin{tabular}[width=\textwidth]{l|ccc|ccc}
\toprule \hline
\multirow{2}{*}{\diagbox{Train Data}{Test Data}} & \multicolumn{3}{c}{SGD} & \multicolumn{3}{|c}{MultiWOZ} \\ \cline{2-7} 
                  & Origin    & AutoAug    & HumanAug & Origin    & AutoAug    & HumanAug    \\ \hline
Origin          & 0.871$_{\pm 0.004}$          & 0.857$_{\pm 0.004}$          & 0.857$_{\pm 0.004}$          & \textbf{0.839$_{\pm 0.001}$} & 0.810$_{\pm 0.003}$          & 0.810$_{\pm 0.003}$          \\
Origin+Syn2\%   & 0.879$_{\pm 0.001}$          & 0.866$_{\pm 0.001}$          & 0.866$_{\pm 0.001}$          & 0.833$_{\pm 0.001}$          & 0.799$_{\pm 0.004}$          & 0.799$_{\pm 0.004}$          \\
Origin+Syn100\% & 0.876$_{\pm 0.001}$          & 0.866$_{\pm 0.001}$          & 0.866$_{\pm 0.001}$          & 0.835$_{\pm 0.002}$          & 0.803$_{\pm 0.003}$          & 0.803$_{\pm 0.003}$          \\
AutoAug             & 0.877$_{\pm 0.006}$          & 0.874$_{\pm 0.005}$          & 0.872$_{\pm 0.005}$          & 0.828$_{\pm 0.005}$          & 0.845$_{\pm 0.006}$          & 0.844$_{\pm 0.007}$          \\
AutoAug+Syn2\%      & 0.879$_{\pm 0.003}$          & 0.878$_{\pm 0.002}$          & 0.876$_{\pm 0.002}$          & 0.826$_{\pm 0.001}$          & 0.860$_{\pm 0.002}$          & 0.859$_{\pm 0.002}$          \\
AutoAug+Syn100\%    & \textbf{0.885$_{\pm 0.004}$} & \textbf{0.886$_{\pm 0.004}$} & \textbf{0.882$_{\pm 0.004}$} & 0.830$_{\pm 0.004}$          & \textbf{0.870$_{\pm 0.005}$} & \textbf{0.870$_{\pm 0.005}$} \\ \hline \bottomrule
\end{tabular}
\caption{The accuracy of the name entity prediction for the whole test set.}
\label{table_name_full}
\end{table*}

\begin{table*}[h]
\centering
\small
\begin{tabular}[width=\textwidth]{l|ccc|ccc}
\toprule \hline
\multirow{2}{*}{\diagbox{Train Data}{Test Data}} & \multicolumn{3}{c}{SGD} & \multicolumn{3}{|c}{MultiWOZ} \\ \cline{2-7} 
                  & Origin    & AutoAug    & HumanAug & Origin    & AutoAug    & HumanAug    \\ \hline
Origin          & 0.369$_{\pm 0.004}$          & 0.131$_{\pm 0.004}$          & 0.101$_{\pm 0.008}$          & \textbf{0.365$_{\pm 0.009}$}     & 0.264$_{\pm 0.015}$          & 0.269$_{\pm 0.011}$          \\
Origin+Syn2\%   & \textbf{0.383$_{\pm 0.008}$} & 0.150$_{\pm 0.015}$          & 0.126$_{\pm 0.011}$          & 0.352$_{\pm 0.007}$              & 0.138$_{\pm 0.033}$          & 0.142$_{\pm 0.039}$          \\
Origin+Syn100\% & 0.372$_{\pm 0.008}$          & 0.190$_{\pm 0.012}$          & 0.160$_{\pm 0.010}$          & 0.365$_{\pm 0.006}$              & 0.197$_{\pm 0.040}$          & 0.192$_{\pm 0.035}$          \\
AutoAug             & 0.358$_{\pm 0.004}$          & 0.303$_{\pm 0.005}$          & 0.238$_{\pm 0.005}$          & 0.338$_{\pm 0.005}$          & 0.419$_{\pm 0.001}$          & 0.415$_{\pm 0.007}$          \\
AutoAug+Syn2\%      & 0.377$_{\pm 0.005}$          & 0.331$_{\pm 0.008}$          & 0.268$_{\pm 0.017}$          & 0.338$_{\pm 0.015}$          & 0.479$_{\pm 0.005}$          & 0.469$_{\pm 0.011}$          \\
AutoAug+Syn100\%    & 0.379$_{\pm 0.015}$          & \textbf{0.351$_{\pm 0.001}$} & \textbf{0.286$_{\pm 0.009}$} & 0.349$_{\pm 0.020}$          & \textbf{0.479$_{\pm 0.008}$} & \textbf{0.481$_{\pm 0.011}$}\\ \hline \bottomrule
\end{tabular}
\caption{Joint goal accuracy evaluated on only the augmented turns. }
\label{table_dst_aug}
\end{table*}

\newpage
\section{Interface of Human Paraphrasing}
\label{sec:appendix}

\begin{figure}[h]
\centering
\includegraphics[width=\textwidth]{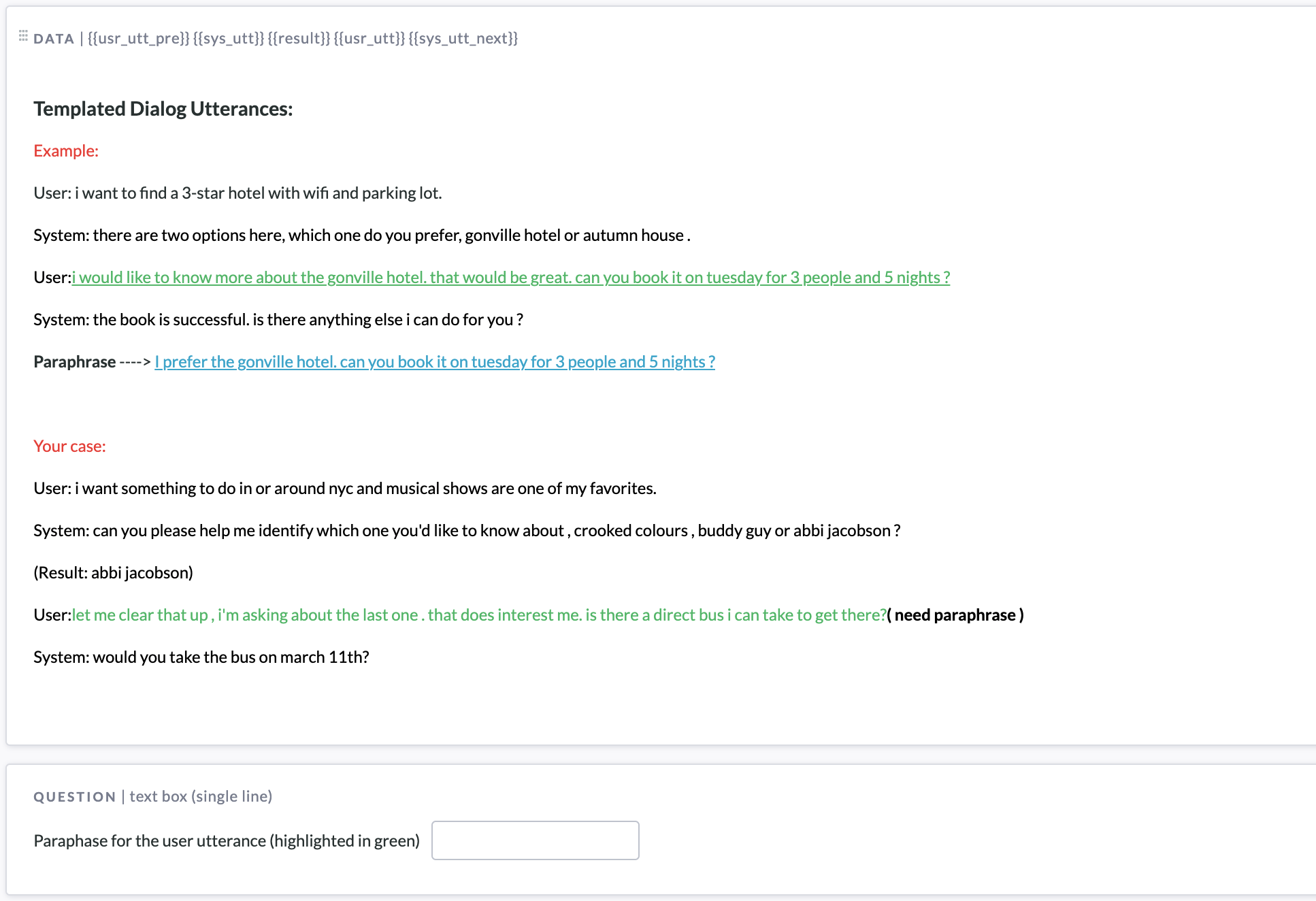}
\caption{Interface to collect human paraphrasing data.
}
\label{fig_human}
\end{figure}

\newpage
\section{Guidelines of Human Paraphrasing}
\begin{figure}[h]
\centering
\includegraphics[width=\textwidth]{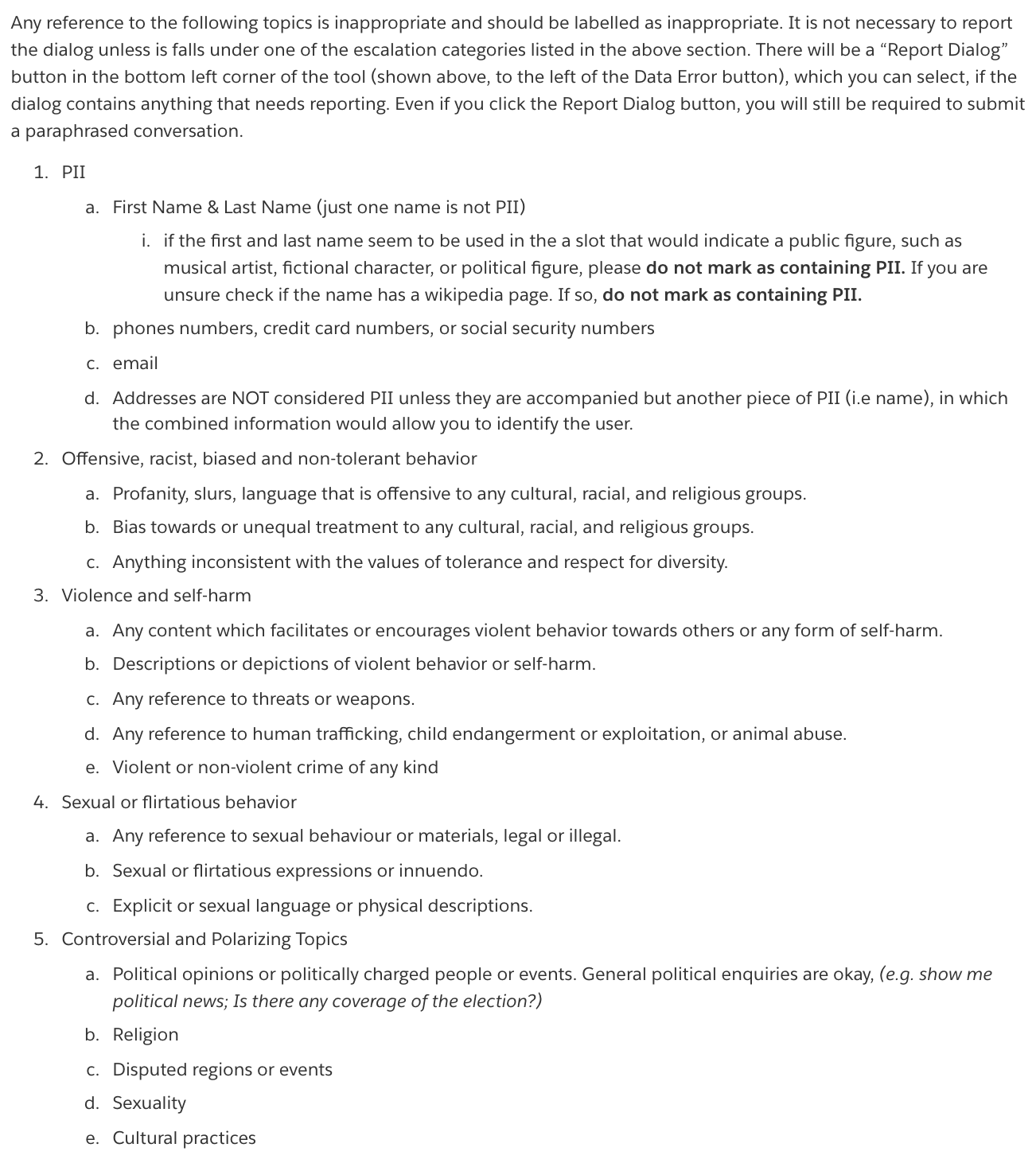}
\caption{Description of sensitive topics.
}
\label{fig_ethical}
\end{figure}

\end{document}